\documentclass[a4paper,amsmath,amssymb]{jpconf}
\usepackage{graphicx}
\usepackage{amsmath, amsthm, amssymb}
\begin{document}
\title{Stochastic gradient method with accelerated stochastic dynamics}

\author{Masayuki Ohzeki}

\address{Department of Systems Science, Graduate School of Informatics, Kyoto University, 36-1 Yoshida Hon-machi, Sakyo-ku, Kyoto, 606-8501, Japan
}

\ead{mohzeki@i.kyoto-u.ac.jp}

\begin{abstract}
In this paper, we propose a novel technique to implement stochastic gradient methods, which are beneficial for learning from large datasets, through accelerated stochastic dynamics. 
A stochastic gradient method is based on mini-batch learning for reducing the computational cost when the amount of data is large. 
The stochasticity of the gradient can be mitigated by the injection of Gaussian noise, which yields the stochastic Langevin gradient method; this method can be used for Bayesian posterior sampling. 
However, the performance of the stochastic Langevin gradient method depends on the mixing rate of the stochastic dynamics. 
In this study, we propose violating the detailed balance condition to enhance the mixing rate. 
Recent studies have revealed that violating the detailed balance condition accelerates the convergence to a stationary state and reduces the correlation time between the samplings. 
We implement this violation of the detailed balance condition in the stochastic gradient Langevin method and test our method for a simple model to demonstrate its performance.
\end{abstract}
\section{Introduction}
Since massive amounts of data can be acquired from various sources, the importance of the so-called big-data analysis is rapidly increasing. 
In particular, extracting relevant parameters that facilitate describing the acquired data is critical for analyzing high-dimensional data.
For high-dimensional data analysis, Boltzmann machine learning can be used to extract parameters that describe the structure of given data \cite{Ackley1985}.
Boltzmann machine learning has proven to be effective and has stimulated increasing interest in deep learning \cite{Hinton2006,Hinton2006sci,RG2014,Ohzeki2015}.
We can here implement the $L_1$-norm regularization to determine the sparse parameters for characterizing given data \cite{Ohzeki2015} and and a type of the relaxation method to the $L_0$ norm \cite{Aurelien2014,Yamanaka2015}.
Although Boltzmann machine learning has been successfully implemented to extract features from high-dimensional data, the computational cost is usually expensive for high-precision estimations. 
The crucial reason for the high computational cost is the expectation value in the Gibbs-Boltzmann distribution. 
In order to mitigate the computational cost, some approximations such as the mean-field analysis, belief propagation, and their improvements can be adopted \cite{Sessak2009,Cocco2011,Cocco2012,Ricci2012,Yasuda2013,Raymond2013,Ohzeki2013}.
In addition, increase of the amount of the data also increases the computational cost for evaluating the update rules in the learning algorithms.

In this study, we develop a technique to mitigate the computational cost of learning from large amount of high-dimensional data. In the middle of the $20$th century, Robbins and Monro proposed a stochastic gradient method, which utilized a mini-batch dataset for gradient computation to update the tentative estimations at each iteration in the learning process \cite{Robbins1951}.
The stochasticity of the gradient will be averaged out because the entire dataset is used by the end of the learning process. 
Then, the learning rate is gradually decreased to ensure convergence to a local maximum of the likelihood function. 
In order to mitigate the stochasticity of the gradient, Welling and Teh proposed an improved version of the stochastic gradient method by introducing Gaussian noise at each iteration such as by using the Langevin dynamics \cite{Welling2011}.
At the first stage of the method, the stochastic gradient noise dominates the injected noise in the learning process. 
On the other hand, at the last stage, the injected noise equilibrates the system to converge to a stationary state.

For enhancing the precision of the learning, it is beneficial to accelerate the convergence to a stationary state in the stochastic gradient Langevin method. Considering the recent advancement in stochastic dynamics in the field of nonequilibrium statistical physics, a novel technique to accelerate the convergence to the desired stationary distribution has been proposed, which performs the violation of the detailed balance condition (vDBC). 
The detailed balance condition (DBC) is typically satisfied in stochastic dynamics simulated by the Langevin equation, the master equation, and the Fokker-Planck equation. 
One such representative method is the Markov-chain Monte-Carlo method. 
The condition assures convergence to a stationary distribution and facilitates the construction of the stochastic rule as in the transition matrix. 
However, the DBC is just a sufficient condition for the convergence and thus is not necessarily to be satisfied. 
Hence, we may violate the DBC to attain faster convergence to the stationary distribution. 
One of the techniques that do not hold the DBC is the Suwa-Todo method \cite{Suwa2010}.
In this method, the elements of the transfer matrix are optimized in the master equation simulated in the Markov-Chain Monte-Carlo method, while ignoring the DBC. Another method is the skewed DBC, which utilizes double stochastic rules for acceleration while violating the DBC \cite{Turitsyn2011}.
The vDBC is characterized by the inherent asymmetry of the transfer matrix at the master equation or the Fokker-Planck equation level. 
This asymmetry leads to an eigenvalue shift as compared to the case when the DBC is satisfied and results in the acceleration of the convergence to a stationary state \cite{Ichiki2013}.
From a different point of view, the vDBC can be captured as rare-event sampling, which produces an alternative pathway to the stationary state \cite{Ichiki2015}.
A simple implementation of the vDBC has been performed in the Langevin dynamics especially for a system with continuous-valued degrees of freedom \cite{Ohzeki2015Langevin,Ohzeki2015proc}.
The vDBC can be recast as the resultant feature of the optimization of stochastic dynamics from a point of view of the variational principle \cite{Takahashi2015}. 
Although several innovative studies to enhance the sampling efficacy and accelerate the convergence to a predetermined distribution have been extensively reported  \cite{Swendsen1987,Hukushima1996,Neal2001,Ohzeki2010a,Ohzeki2010b,Ohzeki2011proc,Ohzeki2011proc2,Ohzeki2012d}, the concept of vDBC is much easier to implement. In this study, we aim to implement the concept of the vDBC in the stochastic gradient Langevin method and verify its performance.

The remainder of the paper is organized as follows: 
First, we briefly review the stochastic gradient Langevin method. 
In the next section, we introduce the scheme of the accelerated stochastic dynamics through the vDBC. 
In the fourth section, we formulate the application of the accelerated stochastic dynamics to the stochastic gradient Langevin method. 
We test our scheme, in the following section, for the simplest learning model, which is a mixture of the Gaussian distribution, to confirm the efficacy of our method. 
The last section is devoted for discussion on the future direction of our study.

\section{Stochastic gradient Langevin method}
In this section, we briefly review the stochastic gradient Langevin method. 
The stochastic gradient Langevin method has been invented for Bayesian learning, which captures the uncertainty in the learned parameters and avoids overfitting. 
This method combines the Robbins-Monro method, which optimizes a likelihood function in a stochastic manner, with overdamped Langevin dynamics, which injects noise into the parameter updates. 
Each stochasticity has a different property. 
The former stochasticity is a resultant property to reduce the computational cost for large-scale data. 
In the latter, the noise makes the trajectory of the parameters converge to full posterior distribution rather than just the maximum a posteriori mode.

We consider the learning of the parameters ${\boldsymbol \theta}$ with a prior distribution $p({\boldsymbol \theta})$ and the probability of the datum ${\bf x}$ given the parameter ${\boldsymbol \theta}$, $P({\bf x}|{\boldsymbol \theta})$; namely the posterior distribution $P({\bf x}|{\boldsymbol \theta}) p({\boldsymbol \theta})$. 
The basic task is to find the maximum a posteriori (MAP) parameters ${\boldsymbol \theta}$ by maximizing the following logarithm of the posterior distribution:
\begin{equation}
{\boldsymbol \theta}^* = \arg \max_{{\boldsymbol \theta}}\left\{ \frac{1}{D}\sum_{k=1}^D \log P({\bf x}^{(k)}|{\boldsymbol \theta})p({\boldsymbol \theta})\right\}.
\end{equation}
In the context of the optimization, the prior distribution regularizes the parameters while the likelihood terms constitute the cost function to
be optimized.
In the basic framework of the maximum a posteriori, by taking the derivative of the parameters ${\boldsymbol \theta}$, we obtain the following iteration for seeking the maximum of the posterior distribution.
\begin{equation}
{\boldsymbol \theta}' = {\boldsymbol \theta} + dt \frac{\partial \mathcal{L}({\boldsymbol \theta})}{\partial {\boldsymbol \theta}},
\end{equation}
where 
\begin{eqnarray}
\frac{\partial \mathcal{L}({\boldsymbol \theta})}{\partial {\boldsymbol \theta}} &=& \frac{\partial}{\partial {\boldsymbol \theta}}\log p({\boldsymbol \theta})
+ 
\frac{1}{D}\sum_{k=1}^D \frac{\partial }{\partial {\boldsymbol \theta}} \log P({\bf x}^{(k)}|{\boldsymbol \theta}). \label{grad1}
\end{eqnarray}
The logarithm of the posterior distribution is defined as $\mathcal{L}({\boldsymbol \theta})$.
Increasing the amount of given data results in the computational cost to sum over the gradient of the log-likelihood in the second term. 
To mitigate the problem, a stochastic gradient method has been proposed. 
In this method, the gradient of the log-likelihood function for a randomly chosen subset of the given data is used: 
 \begin{eqnarray}
\frac{\partial \mathcal{L}({\boldsymbol \theta})}{\partial {\boldsymbol \theta}} &=& \frac{\partial}{\partial {\boldsymbol \theta}}\log p({\boldsymbol \theta})
+ 
\frac{1}{d}\sum_{k=1}^d \frac{\partial }{\partial {\boldsymbol \theta}} \log P({\bf x}^{(k)}|{\boldsymbol \theta}). \label{grad2}
\end{eqnarray}
where $d<D$.
The subset of the given data is randomly chosen.
Over a number of iterations, all the data is used and the noise in the gradient caused by using subsets is averaged out. To ensure convergence to a local maximum, the step size is gradually decreased.
The step size $dt$ depends on the step as $dt_i$ and satisfies $\sum_{i}dt_i = \infty$ and $\sum_{t}dt_i^2 < \infty$ \cite{Robbins1951}.
Typically, step size is set as $dt_i = a(b + i)^{-\gamma}$ where $\gamma \in (0.5,1]$.
This is the stochastic gradient method and just an approximation of the MAP estimation (or also applicable to the maximum likelihood estimation).

To avoid the problem that it does not capture parameter uncertainty and can potentially overfit data in the MAP estimation, stochastic noise as shown below can be introduced.
\begin{equation}
{\boldsymbol \theta}' = {\boldsymbol \theta} + dt \frac{\partial \mathcal{L}({\boldsymbol \theta})}{\partial {\boldsymbol \theta}} + \sqrt{2T}dW,
\end{equation}
where $T$ is the temperature representing the strength of the injected noise (usually set as unity) and $d{\bf W}$ denotes the Wiener process, whose order is $O(\sqrt{dt})$, i. e., the Gaussian noise with a vanishing mean and a variance of $dt$.
This simply represents the overdamped Langevin equation in the context of statistical physics. 
The overdamped Langevin equation has a corresponding Fokker-Planck equation, which describes the time-evolution of the distribution function. 
It can be confirmed that the equilibrium distribution can be the posterior distribution. Welling and Teh have proposed the combination of Langevin dynamics with the stochastic gradient method, i.e., the stochastic gradient Langevin method, to generate the posterior distribution for learning from large-scale data \cite{Welling2011}.
The step size is then decreased similar to the stochastic gradient method introduced above. 
By decreasing the step size gradually, the injected noise will become dominant and the effective dynamics will converge to the Langevin equation with the exact gradient. 
A number of the iterations will thus generate the posterior distribution. 
In other words, the performance of the stochastic gradient Langevin method strongly depends on the convergence rate of the Langevin equation to the equilibrium distribution, i.e., the posterior distribution. 
The increase in the convergence speed improves the performance and reduces the computational cost of the method. 
This fact motivates the study of the stochastic gradient Langevin method from a point of view of nonequilibrium statistical physics.

\section{Accelerated stochastic dynamics}
Here, we employ the recent advancement in stochastic dynamics, i.e., the vDBC. 
For the case of the standard Langevin equation, the DBC holds to ensure the convergence to a stationary state. 
However, this is just a sufficient condition and can be violated. 
We modify the standard form of the Langevin equation to accelerate the convergence speed towards a stationary state. 
This is referred as the accelerated stochastic dynamics. 
In the next section, we introduce the accelerated stochastic dynamics with faster convergence to a stationary state.

\subsection{Langevin equation and its corresponding Fokker-Planck equation}
We begin with the $N$-dimensional overdamped Langevin dynamics defined as
\begin{equation}
d{\boldsymbol \theta} = {\bf A}({\boldsymbol \theta})dt + \sqrt{2T}d{\bf W},
\end{equation}
where ${\boldsymbol \theta}$ is the vector representing the $N$-dimensional degrees of freedom (parameters to be estimated for machine learning) and $d{\boldsymbol \theta}$ denotes the infinitesimal change in ${\boldsymbol \theta}$ during $dt$.
Here we assume that the step size $dt$ is homogeneous for the sake of simplicity.
In addition, ${\bf A}({\boldsymbol \theta})$ is a time-independent force vector, $T$ is the temperature, and $d{\bf W}$ denotes the Wiener process for the $N$-dimensional degrees of freedom as stated earlier.
The force usually takes the gradient of the energy as $- \partial E({\boldsymbol \theta})/\partial {\boldsymbol \theta}$ in the context of the statistical physics.
or our problem, the force is denoted as $\partial \log p({\boldsymbol \theta})/\partial {\boldsymbol \theta} + (1/D) \sum_{k=1}^D \partial \log P({\bf x}^{(k)}|{\boldsymbol \theta})/\partial {\boldsymbol \theta}$.
In other words, the energy function is defined as
\begin{equation}
E({\boldsymbol \theta}) = -  \log p({\boldsymbol \theta}) - \frac{1}{D} \sum_{k=1}^D \log P({\bf x}^{(k)}|{\boldsymbol \theta})
\end{equation}
For simplicity, we use the notion of the energy function hereafter. 
In accelerated stochastic dynamics, we do not stick with the standard case. 
In order to accelerate the convergence to a stationary state, we modify the force from its usual form.
The Langevin dynamics can be formulated as the Fokker-Planck equation as follows:
\begin{equation}
\frac{\partial P({\boldsymbol \theta},t)}{\partial t} = - \sum_{k=1}^N \frac{\partial}{\partial \theta_k} J_k({\boldsymbol \theta},t),
\end{equation}
where ${\bf J}({\boldsymbol \theta},t) = (J_1({\boldsymbol \theta},t),J_2({\boldsymbol \theta},t),\cdots,J_N({\boldsymbol \theta},t))$ denotes the probabilistic flow defined as
\begin{equation}
{\bf J}( {\boldsymbol \theta},t) = \left({\bf A}({\boldsymbol \theta}) - T \frac{\partial}{\partial {\boldsymbol \theta}}\right) P({\boldsymbol \theta},t).
\end{equation}
When the system is in any stationary state, the divergence of the probabilistic flow vanishes. 
Therefore, the following equality should be satisfied in the steady state:\begin{equation}
0 = -  \sum_{k=1}^N \frac{\partial}{\partial \theta_k} J_k^{\rm (ss)}( {\boldsymbol \theta}) \label{BCLan},
\end{equation}
where
\begin{equation}
{\bf J}^{\rm (ss)}( {\boldsymbol \theta}) = \left({\bf A}({\boldsymbol \theta}) - T \frac{\partial}{\partial {\boldsymbol \theta}}\right)P^{(\rm ss)}({\boldsymbol \theta}).
\end{equation}
We refer to the above equality as the divergence-free condition. 
This is a balanced condition in the context of the master equation. 
The stationary state should be set for the accomplishment of our aim as follows: \begin{equation}
P^{(\rm ss)}({\boldsymbol \theta}) = \prod_{k=1}^D P({\bf x}^{(k)}|{\boldsymbol \theta}) p({\boldsymbol \theta}),
\end{equation}
where we set $T=1$ (in machine learning, the inverse of temperature is absorbed into the energy definition or the likelihood function and prior distribution).

We can immediately obtain a trivial solution of the divergence-free condition. 
When the probabilistic flow in the steady state does not exist, the divergence-free condition is satisfied. 
\begin{equation}
{\bf J}^{\rm (eq)}({\boldsymbol \theta}) = {\bf 0},
\end{equation}
where the superscript has been changed from ``{\rm ss}" to ``{\rm eq}".
This stationary state is in particular called the equilibrium state because current does not exist here. 
Then, the force can be defined as
\begin{equation}
{\bf A}({\boldsymbol \theta}) = - \frac{\partial}{\partial {\boldsymbol \theta}}E({\boldsymbol \theta}).\label{eqF}
\end{equation}
If we simulate the Langevin dynamics with the trivial force as shown in Equation (\ref{eqF}), the long-time relaxation yields the equilibrium distribution.
\subsection{Nontrivial solution}
We impose the probabilistic flow in a steady state to obtain the following form:
\begin{equation}
{\bf J}^{\rm (ss)}({\boldsymbol \theta}) = \gamma{\bf B}({\boldsymbol \theta})P^{\rm (ss)}({\boldsymbol \theta}),
\end{equation}
where $\gamma$ is a degree of vDBC.
Following the divergence-free condition, the vector field ${\bf B}({\boldsymbol \theta}) = (B_1({\boldsymbol \theta}),B_2({\boldsymbol \theta}),\cdots, B_N({\boldsymbol \theta}))$ must satisfy
\begin{equation}
0 = \gamma\sum_{k=1}^N \left( \frac{\partial B_k({\boldsymbol \theta})}{\partial \theta_k}
- B_k\frac{\partial E({\boldsymbol \theta})}{\partial \theta_k}
\right)P^{\rm (ss)}({\boldsymbol \theta}) ,
\end{equation}
where we have used the fact that the distribution function is set as $P^{\rm (ss)}({\boldsymbol \theta}) \propto \exp(-E({\boldsymbol \theta}))$ and the subscript for the bracket denotes the element of the vector. 
A trivial solution can be given by 
\begin{equation}
B_k({\boldsymbol \theta}) \propto \exp(E({\bf x})).
\end{equation}
However, we have to be careful about the instability when using the exponential term, which often leads to instability in the integration of the Langevin equality

Let us find an alternative to the force obtained above to remove the instability of the exponential term. 
We set the vector field as
\begin{equation}
B_k({\boldsymbol \theta}) = \left(\frac{\partial E({\boldsymbol \theta})}{\partial \theta_{k-1}} - \frac{\partial E({\boldsymbol \theta})}{\partial \theta_{k+1}}\right),\label{nont1}
\end{equation}
where $\theta_{N+1} = \theta_1$ and $\theta_0 = \theta_N$.
Then the divergence-free condition holds.
The resulting force is given as
\begin{equation}
A_k({\boldsymbol \theta})= -\frac{\partial E({\boldsymbol \theta})}{\partial \theta_k} + \gamma \left(\frac{\partial E({\boldsymbol \theta})}{\partial \theta_{k-1}} - \frac{\partial E({\boldsymbol \theta})}{\partial \theta_{k+1}}\right).\label{DF1}
\end{equation}
This force drives the system while satisfying the divergence-free condition but not the DBC. 
For a two-dimensional case, the above nontrivial force can be implemented as \begin{eqnarray}
A_1({\boldsymbol \theta}) &=& -\frac{\partial E({\boldsymbol \theta})}{\partial \theta_1} - \gamma \frac{\partial E({\boldsymbol \theta})}{\partial \theta_2} \label{nonF21} \\
A_2({\boldsymbol \theta}) &=& -\frac{\partial E({\boldsymbol \theta})}{\partial \theta_2} + \gamma \frac{\partial E({\boldsymbol \theta})}{\partial \theta_1} \label{nonF22}.
\end{eqnarray}
Notice that the nontrivial force can be simply denoted as the combination of the gradient of the energy.

\subsection{Replication of the system}
In addition, we further find a nontrivial solution satisfying the divergence-free condition by considering the replication of the system. Note that this is slightly different from the replica exchange Monte-Carlo simulation \cite{Hukushima1996}.
Let us look at the replicate Fokker-Planck equation 
\begin{equation}
\frac{\partial P(\Theta,t)}{\partial t} = - \sum_{r=1}^R \sum_{k=1}^N \frac{\partial}{\partial  \theta_{rk}} J_{rk}(\Theta,t)\label{DF2},
\end{equation}
where $\Theta$ is a matrix $\Theta = ({\boldsymbol \theta}_1,{\boldsymbol \theta}_2,\cdots,{\boldsymbol \theta}_R)$ containing the $N$-dimensional vector for each system denoted by $r=1,2,\cdots,R$, and ${\bf J}_r(\Theta,t) = (J_{r1}(\Theta,t),J_{r2}(\Theta,t),\cdots, J_{rN}(\Theta,t))$ is the probabilistic flow defined as
\begin{equation}
{\bf J}_r(\Theta,t) = \left({\bf A}_r(\Theta) - T \frac{\partial}{\partial {\boldsymbol \theta}_{r}} \right)P(\Theta,t).
\end{equation}
The gradient indexed by $r$ is considered for each system.
Let us impose the divergence-free condition.
The divergence-free condition for the replicate system can be recast as
\begin{equation}
0 = - \sum_{r=1}^R \sum_{k=1}^N \frac{\partial}{\partial  \theta_{rk}} J_{rk}^{\rm (ss)}(\Theta)\label{BCLan2},
\end{equation}
where
\begin{equation}
{\bf J}^{\rm (ss)}_r(\Theta) = \left({\bf A}_r(\Theta) - T \frac{\partial}{\partial {\boldsymbol \theta}_{r}}\right) \prod_{r=1}^R P^{\rm (ss)}({\boldsymbol \theta}_r).
\end{equation}
We impose the stationary state of the replicate system as the the product of the identical independent distribution. 
We can then determine a nontrivial solution by considering a combination of the forces on the replicate system as 
\begin{eqnarray}
{\bf A}_r(\Theta) &=& - \frac{\partial}{\partial {\boldsymbol \theta}_{r}}E({\boldsymbol \theta}_r) + \gamma \left( \frac{\partial}{\partial {\boldsymbol \theta}_{r+1}}E({\boldsymbol \theta}_{r+1}) - \frac{\partial}{\partial {\boldsymbol \theta}_{r-1}}E({\boldsymbol \theta}_{r-1}) \right),\label{nonF}
\end{eqnarray}
where the boundary index is taken periodically. 
Then the probabilistic flow in the steady state is given as 
\begin{eqnarray}
{\bf J}^{\rm (ss)}_r(\Theta) &=& \gamma \left(\frac{\partial}{\partial {\boldsymbol \theta}_{r+1}}E({\boldsymbol \theta}_{r+1}) - \frac{\partial}{\partial {\boldsymbol \theta}_{r-1}}E({\boldsymbol \theta}_{r-1}) \right)\prod_{r=1}^R P({\boldsymbol \theta}_r).\label{repF}
\end{eqnarray}
We can immediately confirm that the divergence of the above probabilistic flow will vanish because the result of the summation becomes zero. 
However, the probabilistic flow exists even in the stationary state. 
This is quite different from an equilibrium system, where there is no divergence of flow. 
We can confirm that the DBC does not hold in the formulation of the path probability \cite{Ohzeki2015Langevin,Ohzeki2015proc}.
In other words, the remainder of the probabilistic flow is obtained from the vDBC. This is the Ohzeki-Ichiki method. In the original formulation in the literature, a duplicate system was considered, where $R=2$.
Here, we generalize the original formulation into the replicate system for it to be suitable in the application of the stochastic gradient Langevin method.

A series of the previous studies reveal that faster convergence can be assured by a mathematical argument \cite{Ichiki2013} and is understood as the rare-event sampling \cite{Ichiki2015}.
In addition, the Ohzeki-Ichiki method may avoid the critical slowing down that occurs during the phase transition \cite{Ohzeki2015Langevin}.
The method is powerful in this sense and very simple to implement because only the force modification results in remarkable performance. 
The previous study confirms the reduction in the correlation time as well as faster convergence to the stationary state \cite{Ohzeki2015proc}.

Recall the replica exchange Monte Carlo simulation, in which multiple systems with ``different" temperatures are driven simultaneously, while  each realization is exchanged several times.
The technique exhibits outstanding performance with fast convergence to the equilibrium state. 
However, in the above formulation, the replicate system has a common temperature. We can find a nontrivial solution with the inhomogeneous temperature in the replicate system. 
In this study, we restrict ourselves to the case with homogeneous temperature, by considering inhomogeneous temperature to be a part of future work.

\section{Application to the stochastic gradient method}
We are now in a position to apply the Ohzeki-Ichiki method to the stochastic gradient Langevin method. 
We propose two ways to apply this method to the stochastic gradient Langevin method.

\subsection{Ohzeki-Ichiki method}
In the stochastic gradient Langevin method, we compute the gradient of the subset of the given data $(k=1,2,\cdots,d)$.
The indices of the data are randomly chosen.
In the stochastic gradient method, we approximate the force term, i.e., gradient, as follows: 
\begin{equation}
- \frac{\partial}{\partial {\boldsymbol \theta}}E({\boldsymbol \theta}) \approx \frac{\partial}{\partial {\boldsymbol \theta}} \log p({\boldsymbol \theta}) + \frac{1}{d} \sum_{k=1}^d \frac{\partial}{\partial {\boldsymbol \theta}} \log P({\bf x}^{(k)} | {\boldsymbol \theta}).
\end{equation}
A simple way to apply the Ohzeki-Ichiki method is the direct insertion of this approximate gradient into Equation (\ref{DF1}).
This scheme can be directly applied to the code for performing numerical computation without any additional computational cost, which is a remarkable property of the Ohzeki-Ichiki method. 
To enhance precision and improve the computation speed, additional efforts for its performance are usually needed. 
However this is not the case.
In addition, we propose an alternative application of the Ohzeki-Ichiki method by considering the replicate system as shown below.

\subsection{Ohzeki-Ichiki method for replicate system}
In order to obtain more sampling points following the posterior distribution, we may perform parallel computation of the stochastic gradient Langevin method with a smaller batch size. 
This motivation is consistent with the concept of the Ohzeki-Ichiki method, in which we may consider that the replication of the system will generate a nontrivial force. We then distribute the approximate gradient into the replicated system. 
In other words, we apply the gradient determined by a different dataset, i.e., the further divided subset of a given data where the number of subsets is set to be $d_r$ such that $d = \sum_{r=1}d_r$.

We first divide the given subset of $d$ into the replicate system as $d_r$ for each $r$ system.
Then we evaluate the gradient for the replicated system.
We have the value of the gradient for each replicate system as follows:
\begin{equation}
- \frac{\partial}{\partial {\boldsymbol \theta}_r }E({\boldsymbol \theta}_r) \approx \frac{\partial}{\partial {\boldsymbol \theta}} \log p({\boldsymbol \theta}) + \frac{1}{d_r} \sum_{k_r=1}^{d_r} \frac{\partial}{\partial {\boldsymbol \theta}} \log P({\bf x}^{(k_r)} | {\boldsymbol \theta}).
\end{equation}
Then we compute the combination of the gradient conforming the nontrivial force as in Equation (\ref{repF}).
In this sense, the nontrivial force is given by the flow of the gradient computed by different data. 
We call this effect data flow. 
The computation with data flow in our method is equivalent to the stochastic gradient Langevin method because we utilize the same batch size and compute the value of the gradient. 
However, because we replicate the system, we generate more sampling points from the given datasets. 
In this sense, the proposed scheme is very important for sampling following the posterior distribution. 
The average of the posterior distribution can also be evaluated efficiently as in the literature \cite{Ohzeki2015Langevin, Ohzeki2015proc}.
In the next section, we confirm the efficiency of our method.

The most important fact to be again emphasized is simplicity of our method.
In addition, the parallel computation is suitable for performing the Ohzeki-Ichiki method.
\section{Numerical tests}
Let us test the stochastic gradient Langevin method with the Ohzeki-Ichiki method.
Similarly to the first study on the stochastic gradient Langevin method \cite{Welling2011}, we test our algorithm for the Gaussian mixture distribution with tied mean, in which the probability distribution of the data is defined as 
\begin{equation}
P(x | {\boldsymbol \theta}) = \frac{1}{2} \sqrt{\frac{b}{2\pi}} \exp\left( - \frac{1}{2}b\left( x - \theta_1\right)^2\right) + \frac{1}{2} \sqrt{\frac{b}{2\pi}} \exp\left( - \frac{1}{2}b\left( x - \theta_1-\theta_2\right)^2\right)
\end{equation}
where $b$ denotes the precision (the inverse of the variance) and the parameters ${\boldsymbol \theta} = (\theta_1,\theta_2)$ are related to the mean of the Gaussian distribution.
The prior distribution is assumed to be a Gaussian distribution as
\begin{eqnarray}
P({\boldsymbol \theta}) = \left(\frac{a_1a_2}{2\pi}\right) \exp\left( - \frac{1}{2}a_1 \theta_1^2 - \frac{1}{2}a_2 \theta_2^2\right)
\end{eqnarray}
where $a_1$ and $a_2$ denote the precisions for the parameters.
This model consists of a bimodal distribution.
Thus the MAP estimation cannot capture the uncertainty of the parameters.
The posterior distribution can be computed exactly as
\begin{equation}
P({\boldsymbol \theta}|x) = \frac{1}{Z(x)}P(x|{\boldsymbol \theta})P({\boldsymbol \theta}),
\end{equation}
where
\begin{equation}
Z(x) = \sqrt{\frac{\alpha_1}{2\pi}}\exp\left( -\frac{\alpha_1}{2}x^2\right) +\sqrt{\frac{ \alpha_2}{2\pi}}\exp\left( -\frac{\alpha_2}{2}x^2\right)
\end{equation}
and
\begin{eqnarray}
\alpha_1 &=& \frac{a_1b}{a_1+b} \\
\alpha_2 &=& \frac{a_1a_2b}{a_1a_2+(a_1+a_2)b}.
\end{eqnarray}
We generate $D=100$ samples of $x^{(k)}$ from one of the Gaussian distribution at $(\theta_1,\theta_2) = (0,2)$.
The model has another-mode solution as $(\theta_1,\theta_2) = (0,2)$, which has strong negative correlation between the parameters.

\begin{figure}[tb]
\begin{center}
\includegraphics[width=1\textwidth]{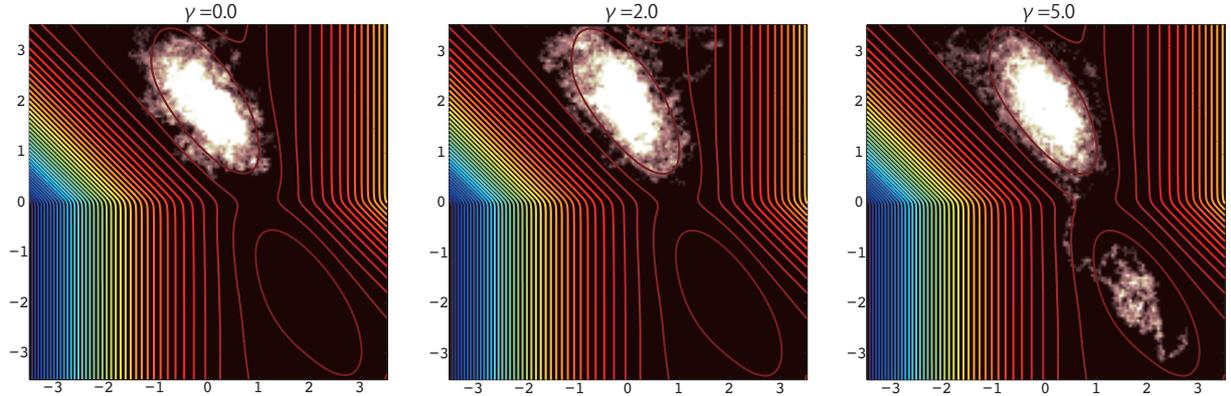}
\end{center}
\caption{ (Colour online)
Density plot and the contour plot of the exact posterior distribution of the batch size $d=1$.  The number of iterations is $\Delta t=10^6$.
The left panel shows the stochastic gradient Langevin method, the centre one describes that with the Ohzeki-Ichiki method with $\gamma=2$, and the right one represents that with $\gamma=5$.
The posterior distribution is plotted by the colour-gradation curves from red (higher values) to blue (lower values).
}
\label{fig1}
\end{figure}

Let us perform the stochastic gradient Langevin method with the Ohzeki-Ichiki method.
The gradient of the log-likelihood function is evaluated as
\begin{eqnarray}\nonumber
\frac{\partial}{\partial { \theta_1}} \log P(x|{\boldsymbol \theta}) &=& \frac{b(x-\theta_1)\exp(-b(x-\theta_1)^2/2) + b(x-\theta_1-\theta_2) \exp(-b(x-\theta_1-\theta_2)^2/2)}{\exp(-b(x-\theta_1)^2/2) + \exp(-b(x-\theta_1-\theta_2)^2/2)} \\
\frac{\partial}{\partial { \theta_2}} \log P(x|{\boldsymbol \theta}) &=& \frac{ b(x-\theta_1-\theta_2) \exp(-b(x-\theta_1-\theta_2)^2/2)}{\exp(-b(x-\theta_1)^2/2) + \exp(-b(x-\theta_1-\theta_2)^2/2)}.
\end{eqnarray}
We set $a_1=a_2=0.1$ and $b=10$, which yield a large barrier between the two modes at $(\theta_1,\theta_2) = (0,2)$ and $(\theta_1,\theta_2)=(2,-2)$.
We iterate the update process for $10^6$ steps.
The step size decreases as $\beta(t+\delta)-{\epsilon}$, where $\epsilon = 0.55$ and $\delta$ and $\beta$ are determined such that step sizes changes from $dt = 0.01$ to $dt = 0.0001$. 
First let us compare the results with the batch size for the stochastic gradient $d=1$.
Under this setting, the original stochastic gradient Langevin method usually fails to generate the sampling points following the posterior distribution in short time.
Here, we demonstrate a more difficult case for the generation of the sampling points following the posterior distribution than the previous study \cite{Welling2011}.
We show the obtained sampling points through a density plot. We compare the stochastic gradient Langevin method with and without the application of the Ohzeki-Ichiki method with $\gamma=2$ and $5$ using the nontrivial force in Equations (\ref{nonF21}) and (\ref{nonF22}) as shown in Fig. \ref{fig1}.
We can confirm that a wide range of the sampling points are obtained by the Ohzeki-Ichiki method.
In particular, the case where $\gamma=5$ exhibits a leap from one mode to the other mode.
In other words, for the case with the application of the Ohzeki-Ichiki method with a larger value of the degree of vDBC estimates more accurate the posterior distribution.

\begin{figure}[tb]
\begin{center}
\includegraphics[width=1\textwidth]{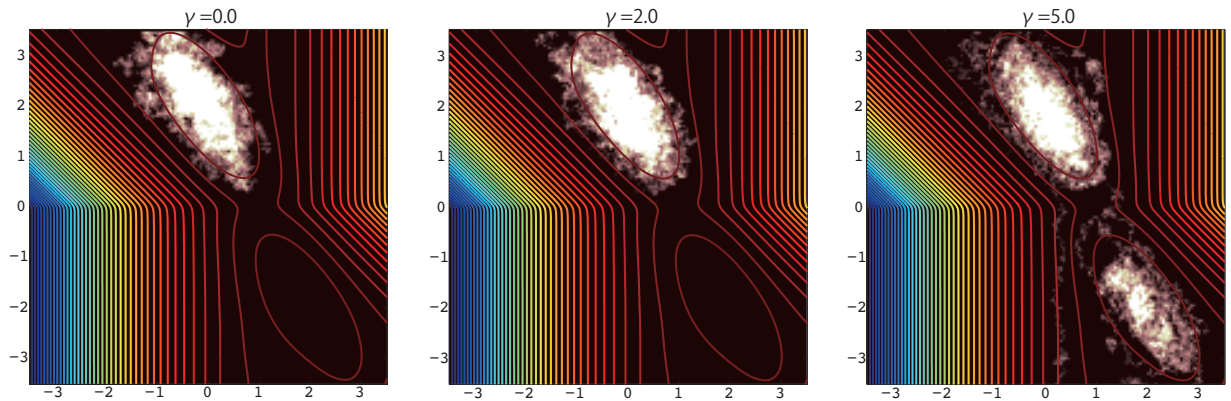}
\end{center}
\caption{ (Colour online)
Density plot and the contour plot of the exact posterior distribution of the batch size $d=10$. The number of iterations is $\Delta t=10^6$.
The same symbols are used in Figure \ref{fig1}.
}
\label{fig2}
\end{figure}

Figure \ref{fig2} shows the results of the stochastic gradient Langevin method and its improved version by the application of the Ohzeki-Ichiki method with the batch size $d=10$.
Similar to the previous case, for $d=1$, we again confirm the efficient sampling by the application of the Ohzeki-Ichiki method.
  
\begin{figure}[tb]
\begin{center}
\includegraphics[width=1\textwidth]{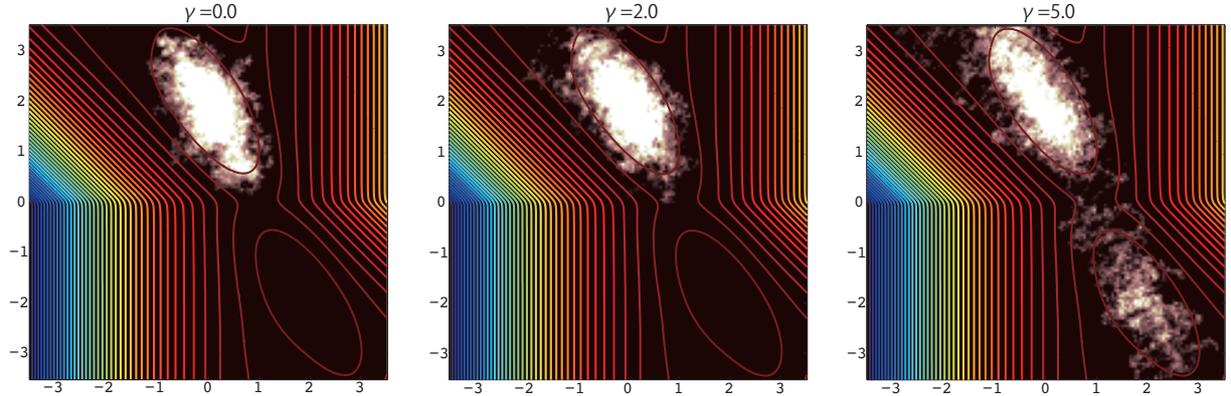}
\end{center}
\caption{ (Colour online)
Density plot and the contour plot of the exact posterior distribution by the data flow of the batch size $d=10$.  The number of iterations is $\Delta t=10^5$.
The same symbols are used in Figure \ref{fig1}.
}
\label{fig3}
\end{figure}

\begin{figure}[tb]
\begin{center}
\includegraphics[width=1\textwidth]{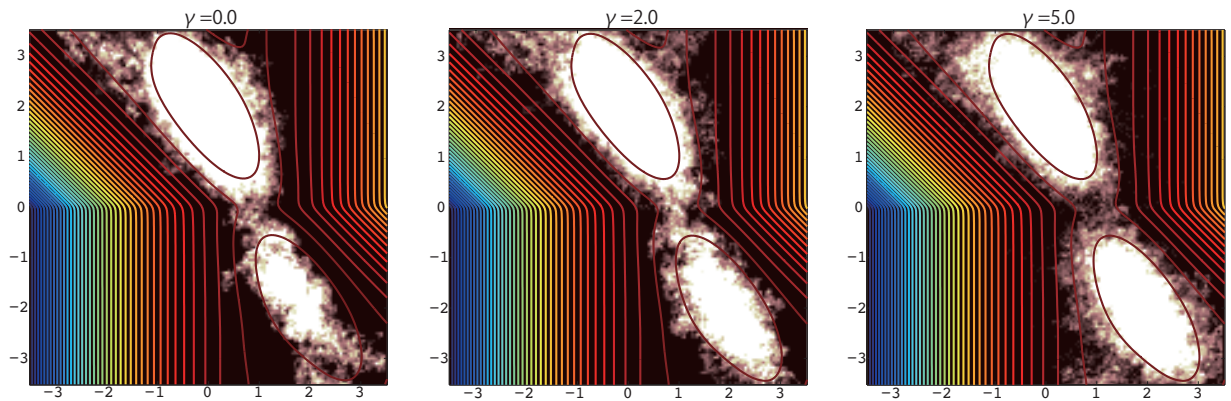}
\end{center}
\caption{ (Colour online)
Density plot and the contour plot of the exact posterior distribution by the data flow of the batch size $d=10$.  The number of iterations is $\Delta t=10^6$.
The same symbols are used in Figure \ref{fig1}.
}
\label{fig4}
\end{figure}

Let us consider another application of the Ohzeki-Ichiki method using the data flow.
We employ the Ohzeki-Ichiki method for the parallel update of the $r=d=10$ replicate system.
For each system, we use one datum randomly chosen for the gradient to update the tentative estimation value. We perform parallel computation of the update equation over $r=d=10$ replicate system.
Then we utilize the data flow coming from the nontrivial force in the Ohzeki-Ichiki method, namely the vDBC.
Figure \ref{fig3} shows the results of the parallel computation of the stochastic gradient Langevin method and those of the data flow induced by the nontrivial force with $\gamma=2$ and $\gamma=5$.
In this computation, we reduce the number of steps to $\Delta t = 10^5$.
Efficient sampling can be performed by using the data flow. 
We should emphasize that here we do not perform summation of the gradient over the subset of the given data. 
We divide the randomly chosen subset into a replicate system and utilize the data for the gradient for each replicate system. 
Then, we introduce the nontrivial force determined from the different data used in each system. 
In other words, we ``reuse" the data for efficient sampling.
We have not yet compared the quantitative performance of the methods.
However one can confirm their qualitative performance from the results shown in the figures.
In addition, we plot the results of the long-time computation up to $\Delta t = 10^6$ by using the data flow shown in Figure \ref{fig4}.
Even for the case without the data flow, the sampling points are found in a relatively wide range.
However increase in the $\gamma$ yields the more efficient sampling following the posterior distribution, as shown in Figure \ref{fig4}.

It may be assumed that the infinitely large nontrivial force causes the system to be in a stationary state immediately. 
This may or may not be true. 
Indeed increase in $\gamma$ makes the system mixed rapidly and accelerates the convergence to a stationary state. However, depending on the value of $\gamma$, the numerical integration of the Langevin dynamics may become worse. To stabilize the computation of the Langevin dynamics, more sophisticated integration methods or the Metropolis-Hasting method may be utilized as in the original proposal of the stochastic gradient Langevin method. 
From mathematical and physical point of view, the Ohzeki-Ichiki method can ensure that a stationary state is reached faster than the case under the DBC. 
Thus, the increase in the degree of vDBC has some tradeoff between performance and instability. 
The optimal value of $\gamma$ will be part of future work. 
The answer may be obtained from the variational principle discussed in a previous study \cite{Takahashi2015}.

\section{Summary}
In this study, we proposed the application of the Ohzeki-Ichiki method in the stochastic gradient Langevin method. 
The Ohzeki-Ichiki method violates the DBC, which is a sufficient condition to ensure convergence to a stationary state, and shows remarkable performance for attaining a stationary state compared to the standard equilibrium case under the DBC. 
We applied this method to the stochastic gradient Langevin method, which approximates the gradient of the log-likelihood function in a stochastic manner and utilizes prior distribution to capture the uncertainty of the leaned parameters. 
The acceleration of the convergence to a stationary state improved the performance of the stochastic gradient Langevin method as expected. 
The range for the sampling points was wider as compared to the original method. 
In addition, we developed a new technique to implement the Ohzeki-Ichiki method based on the idea of the stochastic gradient method. 
We introduced data flow in the replicated system, which is the gradient determined by different data. 
This replication generated more sampling points and the data flow made the replicate system converge to the stationary state faster than the ordinary approach.
These are just the preliminary results of the Ohzeki-Ichiki method for machine learning problems. 
Future study will reveal its significance in this field. 
We again emphasize that the implementation of the Ohzeki-Ichiki method is relatively simpler than other methods to accelerate the convergence to a stationary state. 
This method is simply the addition of the nontrivial force. 
Hence, the proposed method makes advancement in the machine learning process.

\section*{Acknowledgement}
The author would like to thank M. Yasuda, and A. Ichiki for the fruitful discussions. This work is financially supported by the JST-CREST, JSPS KAKENHI Grants No. 25120008 and No. 15H03699, and the Kayamori Foundation of Informational Science Advancement.
\bibliography{jpcs}

\providecommand{\newblock}{}
\begin{thebibliography}{10}
\expandafter\ifx\csname url\endcsname\relax
  \def\url#1{{\tt #1}}\fi
\expandafter\ifx\csname urlprefix\endcsname\relax\def\urlprefix{URL }\fi
\providecommand{\eprint}[2][]{\url{#2}}

\bibitem{Ackley1985}
Ackley D~H, Hinton G~E and Sejnowski T~J 1985 {\em Cognitive Science\/} {\bf 9}
  147--169

\bibitem{Hinton2006}
Hinton G~E, Osindero S and Teh Y~W 2006 {\em Neural Comput.\/} {\bf 18}
  1527--1554

\bibitem{Hinton2006sci}
Hinton G~E and Salakhutdinov R~R 2006 {\em Science\/} {\bf 313} 504--507

\bibitem{RG2014}
Pankaj M and David J~S 2014  {\bf stat.ML/1410.3831}

\bibitem{Ohzeki2015}
Ohzeki M 2015 {\em Journal of the Physical Society of Japan\/} {\bf 84} 054801

\bibitem{Aurelien2014}
Decelle A and Ricci-Tersenghi F 2014 {\em Phys. Rev. Lett.\/} {\bf 112}(7)
  070603

\bibitem{Yamanaka2015}
Yamanaka S, Ohzeki M and Decelle A 2015 {\em Journal of the Physical Society of
  Japan\/} {\bf 84} 024801

\bibitem{Sessak2009}
Sessak V and Monasson R 2009 {\em Journal of Physics A: Mathematical and
  Theoretical\/} {\bf 42} 055001

\bibitem{Cocco2011}
Cocco S and Monasson R 2011 {\em Phys. Rev. Lett.\/} {\bf 106}(9) 090601

\bibitem{Cocco2012}
Cocco S and Monasson R 2012 {\em Journal of Statistical Physics\/} {\bf 147}
  252--314

\bibitem{Ricci2012}
Ricci-Tersenghi F 2012 {\em Journal of Statistical Mechanics: Theory and
  Experiment\/} {\bf 2012} P08015

\bibitem{Yasuda2013}
Yasuda M and Tanaka K 2013 {\em Phys. Rev. E\/} {\bf 87}(1) 012134

\bibitem{Raymond2013}
Raymond J and Ricci-Tersenghi F 2013 {\em Phys. Rev. E\/} {\bf 87}(5) 052111

\bibitem{Ohzeki2013}
Ohzeki M 2013 {\em Journal of Physics: Conference Series\/} {\bf 473} 012005

\bibitem{Robbins1951}
Robbins H and Monro S 1951 {\em Ann. Math. Statist.\/} {\bf 22} 400--407

\bibitem{Welling2011}
Welling M and Teh Y~W 2011 {\em Proceedings of the International Conference on
  Machine Learning\/}

\bibitem{Suwa2010}
Suwa H and Todo S 2010 {\em Phys. Rev. Lett.\/} {\bf 105}(12) 120603

\bibitem{Turitsyn2011}
Turitsyn K~S, Chertkov M and Vucelja M 2011 {\em Physica D: Nonlinear
  Phenomena\/} {\bf 240} 410 -- 414

\bibitem{Ichiki2013}
Ichiki A and Ohzeki M 2013 {\em Phys. Rev. E\/} {\bf 88}(2) 020101

\bibitem{Ichiki2015}
Ichiki A and Ohzeki M 2015 {\em Phys. Rev. E\/} {\bf 91}(6) 062105

\bibitem{Ohzeki2015Langevin}
Ohzeki M and Ichiki A 2015 {\em Phys. Rev. E\/} {\bf 92}(1) 012105

\bibitem{Ohzeki2015proc}
Ohzeki M and Ichiki A 2015 {\em Journal of Physics: Conference Series\/} {\bf
  638} 012003

\bibitem{Takahashi2015}
{Takahashi} K and {Ohzeki} M 2015 {\em ArXiv e-prints\/} (\textit{Preprint}
  \eprint{1509.08212})

\bibitem{Swendsen1987}
Swendsen R~H and Wang J~S 1987 {\em Phys. Rev. Lett.\/} {\bf 58}(2) 86--88

\bibitem{Hukushima1996}
Hukushima K and Nemoto K 1996 {\em Journal of the Physical Society of Japan\/}
  {\bf 65} 1604--1608

\bibitem{Neal2001}
Neal R 2001 {\em Statistics and Computing\/} {\bf 11} 125--139

\bibitem{Ohzeki2010a}
Ohzeki M 2010 {\em Phys. Rev. Lett.\/} {\bf 105}(5) 050401

\bibitem{Ohzeki2010b}
Ohzeki M and Nishimori H 2010 {\em J. Phys. Soc. Jpn.\/} {\bf 79}(8) 084003

\bibitem{Ohzeki2011proc}
Ohzeki M and Nishimori H 2011 {\em Journal of Physics: Conference Series\/}
  {\bf 302} 012047

\bibitem{Ohzeki2011proc2}
Ohzeki M and Nishimori H 2011 {\em Physica E: Low-dimensional Systems and
  Nanostructures\/} {\bf 43} 782 -- 785

\bibitem{Ohzeki2012d}
Ohzeki M 2012 {\em Phys. Rev. E\/} {\bf 86}(6) 061110

\end{thebibliography}
\end{document}